\documentclass{article}
\usepackage{spconf}

\usepackage{graphicx}

\usepackage{helvet}
\usepackage{subcaption}

\usepackage{amsmath} 
\usepackage{array}
\usepackage{mdwmath,bm,bbm}
\usepackage{url}
\usepackage{amsmath,amsthm,amsfonts,float,mathtools}
\usepackage{amssymb}
\usepackage{tikz}
\usepackage{caption}
\usepackage{algorithm}
\usepackage{algpseudocode}
\usepackage{pgfplots}
\usepackage{multirow}

\usepackage{subcaption}

\usepackage{enumitem}
\newtheorem{theorem}{Theorem}

\newtheorem{lemma}{Lemma}

\newcommand{\bproof}{ \begin{IEEEproof} }
	\newcommand{\eproof}{ \end{IEEEproof} }
\newcommand{\beqno}{ \begin{equation*} }
\newcommand{\eeqno}{ \end{equation*} }
\newcommand{\beqa}{\begin{eqnarray*} }
	\newcommand{\eeqa}{\end{eqnarray*} }
\newcommand{\beq}{ \begin{equation} }
\newcommand{\eeq}{ \end{equation} }

\renewcommand{\a}{\bm{a}}

\newcommand{\x}{\bm{x}}
\renewcommand{\b}{\bm{b}}
\newcommand{\y}{\bm{y}}

\newcommand{\X}{\bm{X}}
\newcommand{\U}{{\bm{U}}}

\newcommand{\B}{\bm{B}}
\newcommand{\A}{\bm{A}}
\newcommand{\I}{\bm{I}}
\newcommand{\Y}{\bm{Y}}

\newcommand{\xhat}{\bm{\hat{x}}}

\newcommand{\Uhat}{{\bm{\hat{U}}}}

\newcommand{\E}{\mathbb{E}}

\newcommand{\z}{\bm{z}}

\newcommand{\SE}{\mathrm{SE}}
\newcommand{\dist}{\mathrm{dist}}

\newcommand{\Chat}{\bm{\hat{C}}}

\newcommand{\Span}{\mathrm{span}}
\newcommand{\trace}{\mathrm{tr}}

\newcommand{\cM}{\hat{\bm{D}}}  
\newcommand{\cA}{\bm{D}}    
\newcommand{\cH}{\bm{H}}

\newcommand{\uu}{\bm{u}}

\newcommand{\Lambar}{{\bar{\bm\Lambda}}}

\newcommand{\evdeq}{\overset{\mathrm{EVD}}=} 

\newcommand{\Section}[1]{  \vspace{-0.15in} \section{#1}  \vspace{-0.15in}  } 
\newcommand{\Subsection}[1]{ \vspace{-0.2in} \subsection{#1}  \vspace{-0.1in} }   
\renewcommand{\subsubsection}[1]{\noindent {\bf #1. }}

\newcommand{\dd}{\bm{d}}

\newcommand{\bi}{\begin{itemize}}
	\newcommand{\ei}{\end{itemize}}
\newcommand{\Xhat}{\hat{\bm{X}}}

\newcommand{\chg}{\mathrm{chg}}
\newcommand{\chd}{\mathrm{chd}}
\newcommand{\fix}{\mathrm{fix}}
\newcommand{\new}{\mathrm{add}}

\def\x{{\mathbf x}}

\title{Phaseless Subspace Tracking}
\name{Seyedehsara Nayer and Namrata Vaswani}
\address{Iowa State University, Ames, IA, USA.}
\begin{document}
%
\maketitle
\begin{abstract}
This work takes the first steps towards solving the ``phaseless subspace tracking'' (PST) problem. PST involves recovering a time sequence of signals (or images) from phaseless linear projections of each signal under the following structural assumption: the signal sequence is generated from a much lower dimensional subspace (than the signal dimension) and this subspace can change over time, albeit gradually.
It can be simply understood as a dynamic (time-varying subspace) extension of the low-rank phase retrieval problem studied in recent work.%
\end{abstract}
\Section{Introduction}
\label{sec:intro}


The  Phase Retrieval (PR) problem occurs in many applications such as ptychography, crystallography, astronomy. The original PR problem involves recovering an $n$ length signal $\x$ from the magnitudes of its Discrete Fourier Transform (DFT) coefficients. Generalized PR (see \cite{twf} and \cite{lrpr_tsp}) replaces DFT by inner products with any set of measurement vectors, $\a_i$. Thus, the goal is to recover $\x$ from $|\langle\a_i ,\x\rangle|$, $i=1,2, \dots, m$. It is clear that, without extra assumptions, PR will require $m \ge n$.
In recent works, structural assumptions such as sparsity (see \cite{jaganathan2012recovery,shechtman2013gespar,szameit:12:sbs}) or low-rank (see \cite{lrpr_tsp}) have been incorporated into the PR problem in order to reduce the number of measurements $m$ required for exact or accurate recovery. Low-rank has been used in two ways. One is to assume that a single signal re-arranged as a matrix (or a single image)  is itself approximately or exactly low-rank. The goal is to recover this low-rank ``signal'' from its phaseless linear projections \cite{procrustes, butala}. The second is to assume that a time sequence of signals (or vectorized images) together form a matrix that is well modeled as being low-rank. Each signal/image is one column of this matrix. The measurements are phaseless linear projections of each signal or image (each column of the matrix) \cite{lrpr_tsp}. This problem has been referred to as ``Low-Rank Phase Retrieval (LRPR)'' in \cite{lrpr_tsp} where it was first studied.
%

Another way to interpret the LRPR problem is as follows: a time sequence of signals $\x_t$, $t=1,2,\dots,d$, are generated from an unknown low dimensional subspace, i.e., $\x_t = \U \b_t$, where $\U$ is an $n \times r$ basis matrix (tall matrix with mutually orthonormal columns)  with $r \ll n$, and $\b_t$ is an $r \times 1$ coefficients' vector.
For each $\x_t, t=1,2,\dots,d$, we have $m$ phaseless measurements,
$\y_{i,t} = | \langle\a_{i,t}, \x_t\rangle|, i=1,2,... m, t=1,2,... d$. The goal is to either just recover the subspace, $\Span(\U)$, or to recover both $\Span(\U)$ and the coefficients and hence recover the signals $\x_t$'s (equivalently, recover the low-rank matrix $\X = [\x_1, \x_2, \dots, \x_d]$). The former problem can be called ``phaseless PCA'', although the only known way to exactly recover $\Span(\U)$ involves also recovering the $\b_t$'s via iterative alternating minimization (or gradient descent); see the LRPR algorithms of \cite{lrpr_tsp}.
%
%

\subsubsection{Our problem: Dynamic LRPR or phaseless subspace tracking (PST)}
PST can be simply understood as the dynamic (time-varying subspace) extension of LRPR. Thus, instead of the subspace $\Span(\U)$ being fixed, we assume that it can change with time, albeit slowly.
Often, for long data sequences, e.g., long image sequences or videos, if one tries to use a single lower dimensional subspace to represent the entire data sequence, the required subspace dimension may end up being quite large. This can be problematic because it means that the resulting data matrix may not be sufficiently low-rank. 
In such cases, a better model is to assume that the data lies in a low-dimensional subspace that can change over time, albeit gradually.

The most general model for time-varying subspaces allows the subspace to change by a little at each time. However such a model involves too many unknowns. An $r$ dimensional subspace in $n$-dimensional ambient space is fully specified by $nr$ parameters. But the signal $\x_t$ is an $n \times 1$ vector (has only $n$ unknowns).
Thus, allowing the subspacte to change at each time will result in an increase in the number of unknowns  (rather than a decrease which is the purpose of incorporating structure into the PR problem).
%
%
%
A less general model, but one that allows for a reduction in the number of unknowns, is to assume that the true data subspace is piecewise constant with time. This model has been extensively used in robust subspace tracking literature \cite{rrpcp_perf,rrpcp_dynrpca,rrpcp_medrop} where it in fact helps ensure identifiability of the subspaces (in that problem, only one $n$ length measurement vector is available at each time $t$).

Denote the subspace change times by $t_j$ for $j=1,2,\dots,J$ and let $t_0=0$. Thus, we assume that $\x_t = \U_t \b_t$ where $\U_t = \U_{t_j}$ for all $t = t_j, t_j+1, \dots, t_{j+1}-1$. For simplicity, we sometimes misuse notation and use $\U_j$ to denote $\U_{t_j}$.
The goal is to recover the $\x_t$'s from $m$ phaseless measurements at each time, i.e., from $ \y_{i,t} :=  | \langle\a_{i,t}, \x_{t}\rangle|$, $i=1,2,\dots, m$ for each $t=1,2,\dots,d$.
Under this model and assuming ``slow subspace change'' (quantified in Sec. 1.1), the question is when can one solve this problem using a smaller $m$ per signal than what is needed for LRPR? The LRPR work \cite{lrpr_tsp} has already demonstrated that just exploiting the low-rank assumption enables a reduction in the required $m$ compared to simple PR done for each signal $\x_t$ individually.


\subsubsection{Our Contribution}
This work takes the first steps towards solving the phaseless subspace tracking problem described above. Any subspace tracking problem requires two sub-problems to be solved:
\begin{enumerate}[label=(\alph*)]
	\item given an accurate estimate of the previous subspace, detect if the subspace has changed;
	and
	\item given an accurate estimate of the previous subspace, and given that it is known that the subspace {\em has} changed,  estimate the new  subspace using as few measurements (as short a tracking delay) as possible.
\end{enumerate}
We will henceforth refer to (a) as ``phaseless subspace change detection'' and to (b) as ``phaseless PCA with partial subspace knowledge''.
Of course to solve (b) to $\epsilon$ accuracy for any $\epsilon\geq 0$, the resulting algorithm needs to also estimate the subspace projection coefficients as well. An early version of the current work (one that only solves (b) and only under the assumption that the previous subspace is exactly known) will be presented at Asilomar \cite{lrpr_asilomar}.

\Subsection{Notation, assumptions and some definitions}
\label{sec:problem}

\subsubsection{Notation}
%
$\|\cdot \|$ denotes the $l_2$ norm of a vector or the induced $l_2$ norm of a matrix. For other $l_p$ norms, we use $\|\cdot \|_p$.

A matrix with mutually orthonormal columns is referred to as a {\em ``basis matrix''}.
For basis matrices $\Uhat, \U$, the subspace error ($\SE$) between their respective column spans is quantified by
$\SE(\Uhat, \U ) \coloneqq \|\left( \I - \Uhat\Uhat' \right) \U\|$. This measures the sine of the principal angle between the subspaces.
The {\em phase-invariant distance} between two vectors is quantified using 
$
\dist(\z_1, \z_2) :=  \min_{\phi \in [0,2\pi]} \| \z_1 - e^{\sqrt{-1} \phi} \z_2\|.
$
Normalized column-wise phase-invariant recovery error for matrices $\X$ and $\hat\X$, its estimated version, is computed as $
\mathrm{NormErr}(\X,\hat\X):= \frac{\sum_{k=1}^q \dist( \x_k , \xhat_k )^2 }{ \sum_{k=1}^q \|\x_k\|^2 }$. 
For any two integers $i_1,i_2$, the interval $[i_1:i_2]$ denotes the set of integer values $\{i_1, i_1+1,\ldots,i_2\}$ and interval $[i_1:i_2)$ denotes the set $\{i_1, i_1+1,\ldots,i_2-1\}$.

\subsubsection{Assumptions}
We quantify ``slow subspace change'' using the model from \cite{rrpcp_dynrpca}. 
In \cite{rrpcp_dynrpca} and previous work, this has been successfully used to improve outlier tolerance of dynamic robust PCA as compared to its static counterpart.
``Slow subspace change'' \cite{rrpcp_dynrpca} means that $ \U_t = \U_{t_j}:=\U_j \  \forall \  t \in [t_j : \ t_{j+1}) $ (piecewise constant subspaces) and the following hold:
\\ (a) $\SE(\U_{j-1}, \U_{j}) \le \Delta$ with $\Delta$ small,
\\ (b) at each change time, {only} one direction changes, and
\\ (c) $ \min_j (t_{j+1}-t_j)$ is lower bounded.

%

\subsubsection{Definitions}
As mentioned above, at each subspace change time, only one direction changes, while the rest of the subspace remains fixed. Of course at different change times, the changing direction could be different, thus over a long period of time, the entire subspace could change.
To explain this further,
if $\uu_{{j-1},\chg}$
denotes the direction from $\Span(\U_{{j-1}})$ that changes at $t_j$, and $\uu_{j,\chd}$ denotes its changed version, then $\Span(\U_{{j-1}}) = \Span([\U_{{j-1},\fix}, \uu_{{j-1},\chg}])$ and
$ \Span(\U_{j} ) = \Span([\U_{{j-1},\fix}, \uu_{j,\chd}]) $, where $\U_{{j-1},\fix}$ is an
$n \times (r - 1)$ matrix corresponding to the fixed part of the subspace at $t_j$.
Denote the direction that gets added to the subspace at time $t_j$ by $\uu_{j,\new}$. Clearly,
\[\uu_{j,\new}\coloneqq
\frac{(\I - \uu_{{j-1},\chg} \uu_{{j-1},\chg}{}'
	)\uu_{t_j,\chd}}
{\SE(\uu_{{j-1},\chg}, \uu_{j,\chd})}.\]
Also,
$\theta_j := \cos^{-1}
\lvert \uu_{{j-1},\chg}{}' \uu_{j,\chd} \rvert$ is the angle between $\uu_{j-1,\chg}$ and $\uu_{j,\chd}$
and $\uu_{j,del} := \uu_{{j-1},\chg} \sin{ \theta_{j} } - \uu_{j,\new} \cos{\theta_{j}}$ is the direction that gets deleted at $t_j$. 
	
The following facts are immediate from the above: 
(i) $|\sin{\theta_{j}} | = \sin{\theta_{j}} = \SE(\uu_{{j-1},\chg}, \uu_{j,\chd})$,
(ii) $\uu_{j,\chd} = \uu_{j,\new} \sin{\theta_{j}} + \uu_{{j-1},\chg} \cos{\theta_{j}}$ ,
(iii) $\uu_{j,\new}$ is orthogonal to $\U_{{j-1}}$, and
(iv) $\Span(\U_{j}) \subseteq \Span([\U_{{j-1}}, \uu_{j,\new}])$.

Define the sub-matrix $\X_j:=[\x_{t_j}, \x_{t_j+1}, \dots, \x_{t_{j+1}-1}]$, let  $q := \left(t_{j+1}-t_j\right)$ and let 
\[
\E \left[\frac{1}{q} \sum_{t \in [t_j, t_{j+1})} \x_t \x_t{}' \right] \evdeq \U_j \Lambar_j \U_j{}'
\]
denote its eigenvalue decomposition (EVD).
To simplify notation, in the text below we sometimes remove the subscript $j$, e.g., we often use $\theta$ to denote $\theta_j$.

\Section{Solution approach}
In the next two subsections, 
we explain how to solve each of the two sub-problems mentioned above.%

\Subsection{Automatic phaseless subspace change detection}
\label{sec:Auto_Detection}
 Consider the matrices
\beq
\Y_{U}:=\frac{1}{m q} \sum_{i=1}^m \sum_{t\in [t_j:t_{j+1})} \y_{i,t} \a_{i,t} \a_{i,t}{}' , \mbox{ and}
\label{eq:Ydef1}
\eeq
\beq
\tilde{\Y}_{U}:=
(\I - \Uhat_{j-1} \Uhat_{j-1}{}')\ \Y_{U}\ (\I - \Uhat_{j-1} \Uhat_{j-1}{}').
%
\label{eq:Ydef2}
\eeq

To understand our approach simply, suppose that $\Uhat_{j-1}$ is a perfect estimate, i.e., suppose that $\Span(\Uhat_{j-1}) = \Span(\U_{j-1})$. Then, it is not hard to see that
{\small
\begin{align}
\E[\tilde{\Y}_{U}] & =  (\I - \U_{j-1} \U_{j-1}{}') \left[ 2(\U_{j} \Lambar \U_{j}{}')  + \trace(\Lambar)\I\right](\I - \U_{j-1} \U_{j-1}{}') \nonumber \\
& = (2 \sin^2{\theta}\ \lambda_{\min}(\Lambar)\  \uu_{j,\new} \uu_{j,\new}{}' +  \trace(\Lambar)(\I-\U_{j-1} \U_{j-1}{}').
\label{EtildYU}
\end{align}
}
The first equality follows from \cite[Lemma A.1]{wf}, and the second follows using the subspace change assumption.
Observe that the above matrix is orthogonal to  ${\U}_{j-1}$.   Let $\U_{j-1, \perp}$  be {\em a} basis matrix for the subspace orthogonal to $\U_{j-1}$. Since $\U_{j-1}$ has rank $r$, this will be an $n \times (n-r)$ matrix. Since $\uu_{j, \new}$ is orthogonal to $\U_{j-1}$, thus, without loss of generality, we can assume that $\uu_{j, \new}$ is one of the columns of $\U_{j-1, \perp}$. Denote the matrix for the rest of its columns by  $\check{\U}_{j-1,\perp}$. Thus,  $\Span(\U_{j-1, \perp} ) = \Span([\uu_{j, \new}, \check{\U}_{j-1,\perp}])$ and so, using \eqref{EtildYU}, an EVD of $\E[\tilde{\Y}_{U}]$ is $\E[\tilde{\Y}_{U}] \evdeq$
{\small
 \begin{align*}
 [\uu_{j, \new}, \check{\U}_{j-1,\perp}] \ \begin{bmatrix*}
 2\sin^2\theta  \lambda_{\min}(\Lambar)  +\trace(\bar{ \Lambda}) & 0 \\
 0 & \trace(\bar{ \Lambda})I
 \end{bmatrix*} \begin{bmatrix*}[c]
 \uu_{j, \new}{}' \\ \check{\U}_{j-1,\perp}{}'
 \end{bmatrix*}.
 \end{align*}
}
  %
  Clearly the top eigenvector of this matrix is equal to $\uu_{j,\new}$, with the corresponding eigenvalue of $ 2\sin^2{\theta}\ \lambda_{\min}(\Lambar) +  \trace(\Lambar)$, and a gap of $2\sin^2{\theta}\ \lambda_{\min}(\Lambar)$ between first and other eigenvalues. So, by law of large numbers \cite{vershynin}, with high probability (w.h.p.), that the top eigenvector of this matrix will be a good initial estimation of $\uu_{j,\new}$, when $m$ and $q$ are large enough. Currently we are making an intuitive argument, these statements will be made rigorous in follow-up work.

From above, again by law of large numbers and assuming $\hat\U_{j-1}$ is a good estimate of $\U_{j-1}$, when $m$ and $q$ are large, if the subspace has not changed, the first eigenvalue of $\tilde{\Y}_{U}$, ${\lambda}_1(\tilde{\Y}_{U})$, will be close to $\trace(\bar{\Lambda})$ w.h.p.; while if it has changed, it will be close to $2\sin^2\theta \lambda_{\min}(\Lambar) + \trace(\bar{\Lambda})$ w.h.p.. A natural subspace change detection approach thus involves thresholding ${\lambda}_1(\tilde{\Y}_{U})$. 
Thus, $\lambda_1(\tilde{\Y}_{U}) \geq C \trace(\bar{\Lambda})$ can be used as a criterion for detecting the change with $C$ being a value slightly more than one.
Here $\trace(\bar{\Lambda})$ is unknown but notice that for $i=r+1,r+2,\dots,n$  $\lambda_i(\E[\Y_U]) = \trace(\bar\Lambda)$. Thus, w.h.p., when $m$ and $q$ are large enough, $\lambda_n(\Y_U) \approx \trace(\bar\Lambda)$ and so we use $\lambda_n(\Y_U)$ as an estimate of $\trace(\bar\Lambda)$. Algorithm 1 summarizes our approach. 

%

It is clear from the above that the change detection performance improves as $\theta$ increases. This fact is also observed through our experiments (see the ROC curves in Fig. 1).


\Subsection{Phaseless PCA with partial subspace knowledge}
\label{sec:PST}

After detecting the change, the next step of PST algorithm is estimating the current subspace knowing the existence of a change.
Here in order to use the previously estimated subspace $\Uhat_{j-1}$, we construct a bigger subspace matrix $\tilde{\U}_{j} \in \mathbb{R}^{n \times (r+1)} $ which contains a new added column besides $\Uhat_{j-1}$. Similarly,  the number of rows of $\B_j$ are increased by one and $\tilde{\B}_j \in \mathbb{R}^{(r+1)\times q}$ is a relaxed estimation of $\B_j$.
Phaseless PCA with partial subspace knowledge consists of two steps which we explain in the following. 

\subsubsection{Initialization}
\label{initidea}
The initialization is inspired by the previously proposed spectral method which is used in many existing works like~\cite{twf} and extended in \cite{lrpr_tsp}. From the discussion above, when $m$ and $q$ are large enough and $\Span(\Uhat_{j-1})$ is close to $\Span(\U_{j-1})$, using the Davis-Kahan $\sin \theta$ theorem \cite{davis_kahan}, it can be argued that the top eigenvector of $\tilde{\Y}_{U}$ will be a good estimate of $\uu_{j,\new}$. Denote this by $\hat{\uu}_{j,\new}$.
With this, $\tilde{\U}_j = [\Uhat_{j-1} \ \  \hat{\uu}_{j,\new}]$ can be used as the initial estimated subspace.
%
Using an idea similar to the approach of \cite{lrpr_tsp}, the top eigenvector of
\begin{align}
\Y_b = \tilde{\U }_{j}{}' \left(\frac{1}{m}\sum_{i=1}^m y_{i,t} \a_{i,t}\a_{i,t}' \right)\tilde{\U }_{j},
\label{eq:Yb}
\end{align}
denoted $\hat{\tilde{\b_{t}}}$, will be a good estimate of $\tilde{\b_{t}}:= \begin{bmatrix}
\b_{t,1:r-1} \\
b_{t,r} \cos{\theta}\\
b_{t,r} \sin{\theta}
\end{bmatrix}$.

\subsubsection{Main loop}
Main loop is an alternating minimization solution.
Similar to our previous work~\cite{lrpr_tsp}, this part has three steps. At each step one of the three variables is estimated and the other two is assumed to be constant. Using alternating minimization, $\tilde{\U}$ can be obtained by solving 
\[ 
\tilde{\U } =\arg\min_{\U}  \sum_t \| \Chat_t  \y_t - \A_t{}' \U \hat{\tilde{\b}}_t  \|^2, 
\] 
where $\hat{\tilde{\b}}_t \in \mathbb{R}^{r+1}$ is provided by the previous iteration. In the simple situation where just one direction is changing at a time, recovering $\hat{\uu}_{j, \new}$ is enough. This can be obtained by
\[   
\hat{\uu}_{j, \new} =  \arg\min_{\tilde\uu}  \sum_t  \| \Chat_t  \y_t - \A_k{}' \Uhat_0 \hat{\tilde{\b}}_{t, 1:r} - \A_t{}' \tilde\uu \hat{\tilde{\b}}_{t,r+1}  \|^2 ,  
\] 
where $\hat{\tilde{\b}}_{t,1:r}$ and $\hat{\tilde{\b}}_{t,r+1}$ are vectors that contain the first $r$ elements and the last element of $\hat{\tilde{\b}}_t$ respectively.
 Then matrix $\tilde{\U }_j = [\Uhat_{j-1}, \hat{\uu}_{j, \new}]$ will be the estimate of current subspace.
 Rest of the solution is similar to our previous work. Using the new estimate of $\U_j$, $\tilde{\B_j}$ and the phase are updated respectively. To remove the relaxed dimension whenever is needed, singular value decomposition can be used as the last step. Matrix $\Xhat_j$ can be recovered as a byproduct also.
 
 The complete algorithm is summarized in Algorithm~\ref{Alg:Heur}.
This works well when  the column span of $\Uhat_{j-1}$ is a good estimate of the subspace spanned by $\U_{j-1}$. Its final subspace recovery error is lower bounded by $\SE(\Uhat_{j-1}, \U_{j-1})$. To reduce the error beyond this value, at the end of Algorithm~\ref{Alg:Heur},  a few iterations of LRPR-AltMin (the algorithm of \cite{lrpr_tsp}) can be used. Just a few iterations of LRPR-AltMin will significantly reduce the error because this can be interpreted as beginning LRPR-AltMin with a very good initial estimate.


%

\begin{algorithm}[t]
	\caption{PST-detection} 
	\label{Alg:detect}
	\algrenewcommand\algorithmicrequire{\textbf{Input:}}
	\algrenewcommand\algorithmicensure{\textbf{Output:}}
	\begin{algorithmic}[1]
		\State Compute $\Y_{U}$ and $\tilde{\Y}_{U} $ using (\ref{eq:Ydef1}) and (\ref{eq:Ydef2}).
		\State Declare a change if $\lambda_1(\tilde{\Y}_{U}) \geq C \ \lambda_n(\Y_{U})$. 
	\end{algorithmic}
\end{algorithm} 

\begin{algorithm}[h!]
	{
	\caption{PST-PCA}\label{Alg:Heur}
	\algrenewcommand\algorithmicrequire{\textbf{Input:}}
	\algrenewcommand\algorithmicensure{\textbf{Output:}}
	\begin{algorithmic}[1]
		\Require  $y_{i,t}$ and $\a_{i,t}$ for $i \in [1:m],t\in [t_j:t_{j+1})$, $\Uhat_{j-1}, r, n$
		\State \textbf{Initialization}:
		\State Compute $\tilde{\Y}_{U}$ as described in (\ref{eq:Ydef2}) and compute $\hat{\uu}_{\new}$ as its top eigenvector.
		\State Set $\tilde{\U} = \left[ \Uhat_{j-1}, \hat{\uu}_{\new}  \right] $.
		\For{$t \in \left[t_j, t_{j+1}\right)$}
		\State Compute $\Y_b$ as described in (\ref{eq:Yb}).
		\State Set $\hat{\tilde{\b}}_{t}$ as top eigenvector of $\Y_b$
		and scale it by $\sqrt{\frac{\sum_i y_{i,t}^2}{m}}$.
		\EndFor
		\For{ $\tau \in [0:T_{\max})$}
		\For{$t \in \left[t_j, t_{j+1}\right)$}
		\State set $\Chat_{t} = \mbox{phase}(\tilde{\U} \tilde{\b}_{t})$.
		\State Set $\hat{\tilde{\b}} = \begin{bmatrix}  \hat{\tilde{\b}}_{[1:r],t} \\ \hat{\tilde{\b}}_{r+1,t}\end{bmatrix}$.
		\State Set $\dd_t = \Chat_{t}  \y_{t}  -  \A'_t \Uhat_{j-1}\hat{\tilde{\b}}_{[1:r],t}$
		\EndFor
		\State Compute $\hat{\uu}_{\new}$ as
		\[ \left(\sum_{t \in \left[t_j:t_{j+1}\right)} (\hat{\tilde{\b}}_{r+1,t})^2 \A_t \A_t'\right)^{-1} \sum_{t\in [t_j:t_{j+1})} \hat{\tilde{\b}}_{r+1,t} \A_t \dd_{t} .\]
		\State Set $\tilde{\U} = \left[ \Uhat_{j-1}, \hat{\uu}_{\new}  \right] $.
		\For{$t \in \left[t_j, t_{j+1}\right)$}
		\State set $\hat{\tilde{\b}}_{t} = \mbox{argmin}_{\b} \|\Chat_{t} \y_t -  \A_t' \tilde{\U} \b\|$
		\EndFor
		\EndFor
		\State Compute $\tilde{\X}_{j} = \tilde{\U} \hat{\tilde{\B}}$.
		\Ensure Set $\Uhat_{j}$ as top $r$ eigenvectors of $\tilde{\X}_{j}$, and $\hat{\B}_{j} = \Uhat_{j}' \tilde{\X}_{j}$.
	\end{algorithmic}
}
\end{algorithm} 


\Section{Numerical Experiments}
\label{sec:Experiments}
\label{sec:typestyle}



\subsubsection{Phaseless subspace change detection}
\begin{figure}[h!]
\centering
		\includegraphics[height=3cm,width=0.4\textwidth]{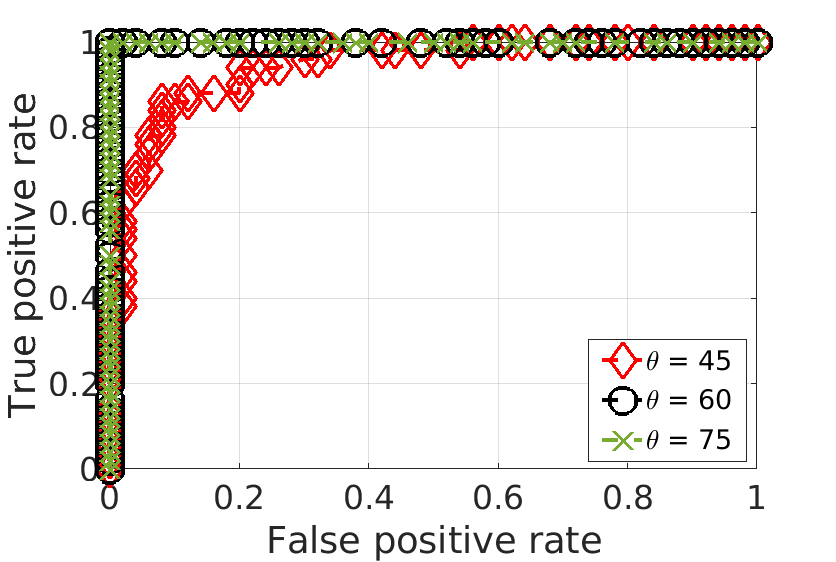}
\vspace{-0.1in}
	\caption{ \label{Fig:ROC} \footnotesize{ROC curve when $\SE(\Uhat_0, \U_0) \approxeq 10^{-4}$}. ``True positives'' refers to probability of correct detection; ``False positives'' refers to false alarm probability.}
\end{figure}

To evaluate the detection performance of Algorithm \ref{Alg:detect}, we plot the receiver operating characteristic (ROC) curve by varying the constant $C$. 50 runs of two sets of data are generated, one without the change and one with the change. For each value of $C$, we  compute the Monte Carlo estimate of the probability of correct detection by using the dataset with change, and we estimate the false alarm probability by using the dataset without change. We then plot the detection probability on the y-axis and the false alarm on the x-axis for various values of $C$.
We show the plots in Fig.~\ref{Fig:ROC} for various values of $\theta$. Settings for this experiment are $n=1000$, $r=10$, $m=850$, $q=750$, $\theta = 30, 45, 60, 75$ degrees and values of $C$ varying between $0$ and $3$, with intervals of $10^{-4}$. Also $\Uhat_0$ was generated so that $\SE(\Uhat_0, \U_0) \approxeq 10^{-4}$.

%

\begin{table}[h!]
	\centering
	\scalebox{0.8}{
\begin{tabular}{||c|ccc|ccc||}
			\hline\hline
			  & & $\SE \approx 10^{-4}$ & &  & $\SE  \approx  10^{-6}$& \\ 
			\hline\hline
			 m& $q$=400 & $q$=500 & $q$=600 & $q$=400 & $q$=500 & $q$=600\\
			\hline
			450		&0.84   &   0.92 	&	0.98  &0.92 &0.98 &	 1.00 \\
			\hline
			550		&1.00	&   1.00    & 	1.00  &1.00 &1.00 &  1.00\\
			\hline
			650		&1.00  	& 	1.00    &   1.00  &0.98	&1.00 &  1.00\\
			\hline\hline
	\end{tabular}}
\vspace{-0.1in}
	\caption{\footnotesize{Success probabilities for two values of $\SE(\hat{\U}_0, \U_0)$. Cases with error of $\U $ less than $1.5 \times \SE(\hat{\U}_0, \U_0)$ are considered successful. PST loop is broken when the subspace error of recovered $\U$ is less than $1.5 \times \SE(\hat{\U}_0, \U_0)$, or when the difference of estimated values of $\U$ between two successive iterations is less than $10^{-9}$.}
	}
	\label{table:1}
\end{table}

	\begin{figure}[t!]
	\centering
	\begin{tikzpicture}[scale=0.5]
	\begin{semilogyaxis}[
	legend style={at={(0.64,0.6)},anchor=north},
	xlabel = \Large{Time taken (seconds)},
	ylabel = \Large{Normalized error of $\hat{\mathbf{X}}$},
	y label style={at={(-0.04,0.5)},anchor=north},
	x label style={at={(0.4,0.0)},anchor=north},
	yscale= 1,
	xscale=1.2,
	xmin = 0,
	xmax = 1150,
	ymin = e-5,
	ymax = 2.5,
	grid,
	] \addplot[line width=0.002cm, thick, mark = square, mark size=6, color = blue] table[x=Tn_700_30,y=Xn_700_30]{FrstExpmNew_n_1000_Xfix.txt} ;
	\addlegendentry{PST-PCA-LRPR};
	\addplot[line width=0.002cm, thick, mark = o, mark size=7, color = red] table[x=To_700_30,y=Xo_700_30]{FrstExpmOld_n_1000_Xfix.txt} ;
	\addlegendentry{LRPR-AltMin};
	\addplot[line width=0.002cm, mark = +,thick,  mark size=7, color = violet]
	table[x=Tw_700_30,y=Xw_700_30]{FrstExpmTWF_n_1000_Xfix.txt} ;
	\addlegendentry{TWF};
	\end{semilogyaxis}
	\end{tikzpicture}
\vspace{-0.1in}
	\caption{\label{fig:PSTLRPR}\footnotesize{Normalized recovery error of $\X$, $\mathrm{NormErr}(\X,\hat\X)$.}}
	\label{fig:result}
\end{figure}
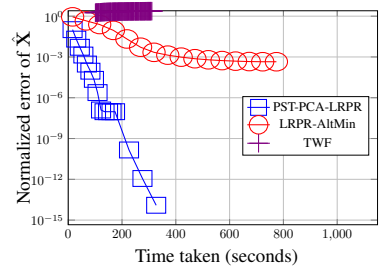

\subsubsection{Phaseless PCA with partial subspace knowledge}
%
In our first experiment, we let $n=1000$, $r=10$, $\theta=30$ degrees. For two values of the initial subspace error $\SE(\Uhat_0, \U_0)$, and many values of $q$ and $m$, we implemented PST-PCA and computed the probability of the final recovery error reaching the same level as the initial subspace error. This is displayed in Table~\ref{table:1}. 50 Monte Carlo runs were used. 
In our second experiment, we compare PST-PCA performance with that of LRPR-AltMin (algorithm of \cite{lrpr_tsp}) and with Truncated Wirtinger Flow (TWF) \cite{twf}. TWF is one of the best known singe signal PR algorithms provably requiring only $O(n)$ measurements for exact recovery.
In this experiment, we implemented PST-PCA for 12 iterations followed by LRPR-AltMin for 3 iterations. LRPR-AltMin was implemented for 15 total iterations. We show our results in Fig.~\ref{fig:PSTLRPR} where we plot the normalized phase-invariant recovery error of $\X$ against the time taken in seconds. This is done by computing the recovery error and time taken at the end of each algorithm iteration. As can be seen just PST-PCA  already significantly outperforms both LRPR-AltMin and TWF (just PST-PCA error decreases to $10^{-7}$ while that of LRPR-AltMin saturates at $10^{-3}$ and TWF is even worse).
This experiment used $n=1000$, $q=500$, $r=15$, $m=700$, $\theta = 30$, and $50$ Monte-Carlo repeats; and value of $\U_0$ is corrupted by an additive Gaussian noise so that $\SE(\Uhat_0, \U_0) = \approx 3.5 \times 10^{-4}$.

\clearpage
\bibliographystyle{IEEEbib}
\bibliography{../../bib/tipnewpfmt_kfcsfullpap}
\end{document}